\documentclass[sigconf,nonacm]{acmart}
\IfFileExists{libertine.sty}{}{\usepackage{lmodern}\usepackage[T1]{fontenc}}
\usepackage{booktabs}
\usepackage{multirow}

\newcommand{\codeurl}{https://github.com/richardzhewang/confidently-wrong}
\newcommand{\ptrue}{P(\textsc{True})}

\title{Confidently Wrong: Detecting Hallucinations in Financial Question Answering from LLM Internal States}
\author{Richard Zhe Wang}
\orcid{0009-0006-3504-6430}
\affiliation{\institution{St.\ John Fisher University}\city{Rochester}\state{NY}\country{USA}}
\email{zwang@sjfc.edu}
\email{rzwang.research@gmail.com}

\makeatletter
\def\acm@copyrightmode{15}
\renewcommand{\@copyrightpermission}{%
  This is the author's version of the work. The definitive Version of Record will appear in the
  \emph{7th ACM International Conference on AI in Finance (ICAIF '26)}, November 14--17, 2026,
  Milan, Italy, \url{https://doi.org/10.1145/3858063.3858685}.}
\makeatother

\begin{document}

\begin{abstract}
Large language models (LLMs) in financial applications fail most consequentially
when they are \emph{confidently wrong}. Hedged, uncertain answers invite scrutiny,
whereas confident errors silently degrade downstream decisions without
warning. We ask how reliably such confidently wrong answers, or confident
hallucinations, can be detected from a model's internal activations, and whether those
activations carry information beyond its output-level confidence signals. We train linear
probes on the residual stream and evaluate them on two established numerical question-answering
(QA) benchmarks built from real filings, FinQA and TAT-QA. Behavioral confidence is measured as the
agreement among eight resampled answers to the same question, and probe effectiveness
is compared against baselines, such as token log-probabilities
and the model's own \textsc{True}/\textsc{False} self-assessment of its answer. Our findings show that among
confident answers, those for which all eight resamples agree, 15--23\% are wrong on
FinQA. There the probes have a significant advantage over output-level baselines in detecting
hallucinations, holding 0.68--0.77 AUROC while the best baselines fall to 0.55--0.63,
across Qwen3-8B, Llama-3.1-8B, and
Gemma-2-9B. Our results suggest that probing can be a cost-effective triage mechanism for routing LLM answers to human review and quality control procedures in high-stakes financial applications. Code and data are available at \url{\codeurl}.
\end{abstract}

\maketitle
\emergencystretch=1em

\section{Introduction}\label{sec:intro}

Assistants built on large language models (LLMs) are entering financial analysis and
advisory workflows, answering questions and supporting high-stakes decisions in financial planning, investment, risk management, and market intelligence \citep{wu2023bloomberggpt,li2023llmfinance,nie2024surveyfinllm}. In these settings, a costly failure mode is LLM \textit{hallucination} of the \emph{confidently} wrong kind. If an answer is hedged or carries detectable signatures of uncertainty, such as low token log-probability, the user can discount the answer or the system can route it to external review. Confidently wrong answers, however, pass both automated filters and the human reader's own judgment. Such errors are especially dangerous in financial applications because the mistakes are often costly and irreversible. An equity investment based on a fabricated growth rate or a misread balance-sheet line could cost an institutional investor millions of dollars. Thus, detecting confidently wrong answers is one of the most important safeguards for deployed financial LLM systems.

Two recent lines of work tackle this problem to various degrees.
\citet{simhi2025choke} show that models
hallucinate with high certainty even when they demonstrably possess the correct
knowledge, and that uncertainty-based detectors fail on precisely these cases.
Separately, a maturing literature shows that answer correctness is linearly decodable
from a model's hidden states on fact-recall benchmarks
\citep{azaria2023internal,burns2023dlk,orgad2024llms,aiersilan2026midlayer}, and
probe-based screening operates at production scale in safety cascades
\citep{cunningham2026classifiers,kramar2026gemini}. To our best knowledge, no
prior work has measured, on numerical financial QA and among high-confidence answers where common output-level confidence signals fail, whether signals from the LLM's internal representations detect the errors that those output-level signals miss.

Our investigation focuses on three questions about numerical financial QA. \textit{First}, does an internal probe beat
deployment-cheap output-level signals at detecting wrong numerical answers, especially confidently wrong
ones? \textit{Second}, which probe design is most effective? \textit{Third}, does a single design
generalize across model families? We then translate the detection method into operational
terms, exploring its triage value in routing LLM answers to human review.

We answer these questions via experiments on two numerical financial QA benchmarks, FinQA and TAT-QA. Our design first crosses
\emph{confidence}, measured behaviorally as the agreement of $k{=}8$ resampled answers
to each query, with \emph{correctness}, established by mechanical grading
against gold answers. We then evaluate, within the resulting $2\times2$ grid, whether a probe on the LLM's internal states can
beat the output-level confidence signals that are available at negligible or low marginal
cost in production (henceforth \emph{deployment-cheap}). These signals are token log-probabilities,
which are a free byproduct of decoding, and \ptrue{}, the model's
\textsc{True}/\textsc{False} assessment of its own answer \citep{kadavath2022}, which
costs one extra forward pass. Throughout, \emph{output-level signals} refers to these scores computed from the model's output distribution. We do not analyze the generated text itself with a separate verifier. Self-consistency costs eight generations per query and is
therefore reported as a reference but excluded from the deployable set.

This paper contributes to the literature in the following ways:
\begin{enumerate}
  \item \textbf{Activation probing for numerical financial QA.} We present, to our
  knowledge, the first systematic evaluation of answer-correctness probing on
  numerical financial QA with modern instruction-fine-tuned LLMs, using two benchmarks built from real filings.
  The probe reaches 0.73--0.79 AUROC (area under the
  receiver operating characteristic curve) over all answers. Fact-recall benchmarks report 0.9--1.0
  \citep{aiersilan2026midlayer}. However, those numbers come from different benchmarks, models, labels, and protocols, so the gap cannot be attributed to financial reasoning alone.
  \item \textbf{Probe's advantage in detecting errors in confident answers.} Among confident answers (unanimous across eight resamples), a linear probe on the 2/3-depth residual stream has an
  AUROC of 0.68--0.77, compared to 0.55--0.63 for the best single output-level signal. On FinQA the probe leads by $\Delta =$ +0.151, +0.151, +0.131 AUROC for Qwen3-8B, Llama-3.1-8B, and Gemma-2-9B, with bootstrap intervals that exclude zero, and it leads a supervised combination of the output-level signals by +0.11 to +0.16. A probe read before generation recovers most of this advantage, and conditioning on the answer adds up to +0.04. The advantage is smaller among unsure answers, where output-level signals remain informative.

  \item \textbf{Probe design, robustness, and validity controls.} A mean-pooled probe at 2/3 of the network depth
  generalizes across the three families of LLMs tested (\S\ref{sec:design}). The confident-answer advantage is not an artifact of the
  probe flagging abstentions, and it persists when unit-dependent scale matches are graded as errors (\S\ref{sec:task}). The advantage does not, however, survive task transfer
  (FinQA$\leftrightarrow$TAT-QA costs 0.07 to 0.16 AUROC), so the probing monitor is task-specific. Every effect is read against random-initialization placebos, which place the
  reference AUROC at $\approx$0.63 rather than 0.5, under document-grouped cross-validation with document-clustered bootstrap intervals (\S\ref{sec:eval}).

  \item \textbf{Which errors the probe detects.} A mechanical typing of FinQA errors shows that the probe's advantage is largest for computation errors, in which the model reads the correct operands but applies a wrong operation, and for sign errors. On both types, output-level signals reach at most 0.64 AUROC (\S\ref{sec:errtypes}).

  \item \textbf{Operational value: triage under bounded review.} In human-in-the-loop
  deployments, at a fixed review budget of 20\% of answers (Qwen3-8B on FinQA),
  probe-ranked routing catches 45\% of all errors, against 37\% for \ptrue{}, 38\% for the supervised output combination, and 20\% for
  random routing. The gain is largest for the over-committing models,
  whose output-level signals are the least informative.
\end{enumerate}

\section{Related work}\label{sec:related}

Three lines of prior work are relevant to this research: (1) truthfulness probing of
hidden states, (2) uncertainty signals and their costs, and (3) financial hallucination detection.

\paragraph{Truthfulness probing.} Prior literature shows that LLM hidden states encode answer correctness. Linear probes and unsupervised directions detect lies and errors
\citep{azaria2023internal,marks2023geometry,burns2023dlk} on fact-oriented benchmarks
such as TruthfulQA \citep{lin2022truthfulqa} and HaluEval \citep{li2023halueval}, and
internal states support both detection and steering of hallucination
\citep{li2023iti,chen2024inside}. Moreover, correctness information concentrates in
specific tokens and mid-network layers \citep{orgad2024llms,aiersilan2026midlayer}, and
probes already screen production traffic in deployed safety cascades
\citep{cunningham2026classifiers,kramar2026gemini}. For a broader survey, see \citet{huang2025survey}. Given these findings, we treat the decodability of
correctness as a premise for the rest of the paper, rather than a new claim.
Our contribution is to quantify the effectiveness of probing signals under confidence stratification, against
deployment-fair baselines, and across model families, in numerical financial QA, where errors frequently arise from multi-step numeric reasoning.

\paragraph{Uncertainty signals and their costs.} Prior work has shown that semantic entropy
\citep{kuhn2023semantic,farquhar2024} and sampling-based agreement measures
\citep{wang2023selfconsistency,manakul2023selfcheckgpt} are strong correctness
predictors, but they require multiple generations per query and thus are not practical for most deployments. Semantic-entropy probes
\citep{kossen2024sep} amortize that cost, although they do not match the sampled
measures. \ptrue{} self-assessment \citep{kadavath2022}, by comparison, adds only one
forward pass. \citet{simhi2025choke} argue that high-certainty hallucinations mostly defeat uncertainty-based methods, and propose probing-based mitigation on perturbation-induced factual errors. A follow-up
work by \citet{simhi2025hack} maps hallucinations jointly along certainty and knowledge axes. We take this phenomenon as our starting point and ask whether probing signals detect hallucinations among confident answers in numerical financial QA, and how large their advantage is over deployment-cheap competitors.

\paragraph{Financial hallucination detection.} Existing systems for financial hallucination
detection focus on \textit{retrieval} and verify outputs against retrieved evidence or estimate entropy over samples \citep{fingground2026,fred2025,ecl2025},
and benchmarks abound \citep{zhang2025faith,cho2026kfinhallu,kumar2026hallubench}.
The exception is a mechanistic case study by \citet{mirajkar2025liar} that traces arithmetic-error circuits in GPT-2~XL on ConvFinQA and reports near-perfect transfer of a linear probe between two families of synthetic templated prompts. In contrast, we find that transfer between two real filing-based benchmarks costs 0.07 to 0.16 AUROC (\S\ref{sec:shift}). Beyond that case
study, no prior work evaluates internal-state detection on modern instruction-fine-tuned LLMs against deployment-relevant baselines at scale.
Furthermore, reported progress in hallucination detection is highly sensitive to evaluation and
benchmark-construction artifacts \citep{janiak2025illusion,hussain2026parallax}. We observe the same in the financial domain, where label noise
and split leakage can each meaningfully move the central comparisons
(\S\ref{sec:task}, \S\ref{sec:eval}), and we control for both.

\section{Task, data, and labels}\label{sec:task}

\paragraph{Task \& models.} Each model receives a financial filing excerpt (text plus a
serialized table) and a question requiring it to locate one to three values amid
distractors and compute a derived quantity (a change, ratio, or share of total). The model
reasons step by step via prompted chain-of-thought (CoT) and emits
\texttt{ANSWER: <number>}. Decoding is greedy with at most 400 new tokens, so each question has one canonical
served answer. Numerical QA is also methodologically convenient, as gold answers are mechanically gradable and labeled examples of model error can be
produced at inference cost, without human annotation.

We evaluate three instruction-fine-tuned LLMs from different developers with distinct pretraining
recipes: Qwen3-8B, Llama-3.1-8B-Instruct, and Gemma-2-9B-it. All three run through an identical pipeline in half precision, with one shared system prompt, chat-template
differences normalized, and Qwen3's native reasoning mode disabled for cross-family
comparability.

\paragraph{Data.} We use two established benchmarks in the versions distributed with the FLARE benchmark \citep{xie2023pixiu}. FinQA \citep{chen2021finqa}
comprises questions written by financial analysts over S\&P~500 firms' filing excerpts
(1999--2019) and requires multi-step numeric reasoning over long contexts. We use its training split (6{,}251 questions), which is the largest. TAT-QA
\citep{zhu2021tatqa}, in contrast, poses questions over annual-report tables with
short accompanying text, and its questions are shorter and more extractive. We use the single split of TAT-QA that FLARE distributes (1{,}668 questions).
Question selection requires the
gold answer to be a bare number, and contexts over 2{,}800 tokens are dropped. The length filter uses each model's own tokenizer, so the FinQA pool differs slightly across models. The FinQA pool contains 6{,}105 questions for Qwen3-8B, 6{,}114 for Llama-3.1-8B, and 6{,}110 for Gemma-2-9B (98\% of the split). The TAT-QA pool contains 1{,}138 questions for every model (68\% of the split). No fine-tuning is performed, and every probe is trained and evaluated by cross-validation within these pools (\S\ref{sec:eval}).

\paragraph{Grading as a measurement instrument.} To grade each answer, the final \texttt{ANSWER:} line is first
parsed to a float, with handling for currency symbols, thousands separators, percent signs, and
accounting-style parenthesized negatives. An answer is graded as correct when it matches the gold answer to within a
2\% relative tolerance under at least one of five scale conventions. Formally,
prediction $p$ matches gold answer $g$ iff
\begin{equation}\label{eq:match}
\operatorname{match}(p,g) \;=\;
\mathbb{1}\!\left[\,\min_{s\in S}\ \frac{|\,ps-g\,|}{\max(|g|,\varepsilon)} \,\le\, \tau\right],
\end{equation}
with tolerance $\tau=0.02$, a small constant $\varepsilon$ guarding division by
zero, and a scale set $S$ containing $1$, $100$, $0.01$, $1000$, and $0.001$,
covering the percent-vs-decimal and thousands-vs-millions conventions of
financial text.
Because every $s\in S$ is positive, no rescaling can repair a
sign flip, so sign errors are never forgiven.

The grader itself is validated in two ways. First, an LLM judge (GPT-5.4-mini) instructed with the
same conventions re-graded 120 query/answer pairs from an 800-question Qwen3-8B FinQA pilot (60 graded-correct and 60 graded-incorrect items). The LLM judge agrees
with the grader on 90.0\% and 93.3\% of them, respectively. The residual disagreement
is dominated by convention ambiguities on which the judge itself is inconsistent
across runs. We estimate residual label noise of 4--6\%, balanced between the two
directions.

Second, because the judge shares the grader's scale conventions, we audit every answer that is accepted only through a non-identity scale. On FinQA, 58--61\% of graded-correct answers fall in this group, because the executable gold answer is a decimal fraction whereas models answer in percent. For 96--98\% of the group, the question asks for a percentage or ratio (in nearly all cases the FinQA annotators' own answer string is written in percent), so the rescaled match is an equivalent representation. The remaining matches, mostly thousand-scale, depend on the unit fixed by the context. They cover 0.9--1.4\% of FinQA answers and 2.9--5.1\% of TAT-QA answers. As a bound, we grade all of them as errors, recompute confidence with the same restricted rule, and refit the probe. The probe's lead among confident answers (\S\ref{sec:fullpop}) changes from $+0.151$, $+0.151$, $+0.131$ to $+0.152$, $+0.180$, $+0.130$ on FinQA, and from $+0.06$--$0.12$ to $+0.12$--$0.17$ on TAT-QA. The conclusions therefore do not depend on the unit-dependent matches.

\paragraph{The 2$\times$2.} Each question $x$ contributes two labels: the correctness of the greedy answer $a_0(x)$ under the grader above, and a behavioral confidence (self-consistency) score \citep{wang2023selfconsistency}
\begin{equation}\label{eq:conf}
c(x) \;=\; \frac{1}{k}\sum_{i=1}^{k} \operatorname{match}\big(a_i(x),\, a_0(x)\big),
\qquad k=8,
\end{equation}
where $a_1(x),\dots,a_k(x)$ are independent resamples of the same query (temperature 0.7, top-$p$ 0.8, top-$k$ 20) and the same
grader (Eq.~\ref{eq:match}) defines agreement. Agreement among resampled answers is
a well-established behavioral confidence measure
\citep{wang2023selfconsistency,manakul2023selfcheckgpt,kuhn2023semantic,farquhar2024}. We call answers that are unanimous across all 8 resamples ($c(x)=1$) \emph{confident answers} and the rest \emph{unsure answers}. The looser threshold $c(x)\ge 6/8$ is reported as a sensitivity check.
Self-consistency was run on every wrong answer but only a random subset of correct answers (FinQA: 1{,}500 correct; TAT-QA: 400 correct; all three models) to save compute. Thus, when computing prevalence, precision, recall, and related shares, each correct example is weighted by the inverse of its sampling rate so the numbers match the full answer pool (population-reweighted).

Table~\ref{tab:quadrants} gives the resulting populations. Qwen3-8B is the most accurate (69.9\%) and the most confident (82\% of answers), while Llama-3.1-8B is the least accurate (60.1\%) and the least confident (36\%). The main cell of interest is the confident-wrong (CW) cell. Qwen3-8B has the most confident-wrongs (19\%), followed by Gemma-2-9B (14\%) and Llama-3.1-8B (5\%). Thus, Qwen3-8B and Gemma-2-9B tend to over-commit, whereas Llama-3.1-8B tends to under-commit. On TAT-QA (untabulated), all three models are more accurate, and the confident-wrong share is 4--7\%.

\begin{table}[t]\centering\small
\caption{\textbf{Confidence $\times$ correctness on FinQA.} Reweighted population
shares of the four cells (C/U = confident/unsure, C/W = correct/wrong). Confident = 8/8 self-consistent. $n$ = questions after filtering; Acc = overall accuracy; CW $n$ = raw count of confident-wrong answers.}
\label{tab:quadrants}
\setlength{\tabcolsep}{4pt}
\begin{tabular}{lrrrrrrr}
\toprule
 & $n$ & Acc & CC & \textbf{CW} & CW $n$ & UC & UW\\
\midrule
Qwen3-8B  & 6{,}105 & 69.9\% & 64\% & \textbf{19\%} & 1{,}137 & 6\%  & 11\%\\
Llama-3.1-8B & 6{,}114 & 60.1\% & 30\% & \textbf{5\%}  & 329     & 30\% & 35\%\\
Gemma-2-9B   & 6{,}110 & 68.1\% & 58\% & \textbf{14\%} & 861     & 10\% & 18\%\\
\bottomrule
\end{tabular}
\end{table}

\section{Methods}\label{sec:methods}

\subsection{Signals under comparison}\label{sec:signals}
The \emph{deployment-cheap set} of output-level signals contains two kinds of signal. The first kind comprises three token-log-probability scores that are free byproducts of decoding: the mean log-probability over all generated tokens, and the mean and the minimum log-probability over the final-answer span (tokens from the last \texttt{ANSWER:} marker onward, or the full generation if the marker is absent).

The second kind is \ptrue{} \citep{kadavath2022}, which shows the model its own question and answer in a new prompt and asks whether the answer is correct (\textsc{True}\slash\textsc{False}). The score is the relative logit of \textsc{True} at the first output position:
    \begin{equation}\label{eq:ptrue} \ptrue{}(x) \;=\;
    \sigma\!\big(
    \log\textstyle\sum_{t\in\mathcal{T}} e^{z_t}
    \;-\;
    \log\textstyle\sum_{t\in\mathcal{F}} e^{z_t}
    \big),
    \end{equation}
where $z$ is the logit vector at the first output position, $\mathcal{T}$ and $\mathcal{F}$ are the tokenizer ids for the case variants of \textsc{True} and \textsc{False}, and $\sigma$ is the logistic sigmoid. \ptrue{} costs one additional forward pass over a prompt that repeats the question, the context, and the answer.

Our main detection method, the \emph{probe}, follows the linear probing-classifier
methodology \citep{alain2016probes,belinkov2022probing} and its application to
truthfulness detection in LLMs \citep{azaria2023internal,orgad2024llms}. The probe is
logistic regression on a linear readout of the residual stream, the model's per-token
internal representation, taken at a layer fixed a priori at $2/3$ depth
and mean-pooled over the generated answer's tokens. Specifically,
\begin{equation}\label{eq:probe}
\bar h(x) \;=\; \frac{1}{|T_x|}\sum_{t\in T_x} h^{(\ell)}_t(x),
\qquad
m(x) \;=\; \sigma\!\big(w^{\top}\bar h(x) + b\big),
\end{equation}
where $T_x$ is the set of generated tokens, $h^{(\ell)}_t(x)$ the residual-stream
vector at token $t$ and layer $\ell=\lfloor 2L/3\rfloor$ of an $L$-layer model, and
$(w, b)$ are fit by $\ell_2$-regularized, class-balanced logistic regression against the
binary \textit{correctness} label ($C{=}0.1$; features standardized within training folds).
The linear probing score, $m(x)$, scores each answer for correctness.

In our experiments, the activations are cached by running the model once more over the prompt followed by its generated answer. In a causal decoder, the hidden state at each token depends only on the preceding tokens, so these are the hidden states that the model computes while it generates the answer. A deployed probe therefore reads these hidden states during generation, needs no additional forward pass, and adds one inner product per answer. The probe requires white-box access and is therefore limited to self-hosted models such as ours. The layer and pooling choices are fixed a priori rather than tuned, and \S\ref{sec:design} reports the
layer-by-pooling sweep behind the $2/3$-depth, mean-pooled choice.

We also report three supervised comparison signals trained with the probe's labels and folds. The \emph{output combination} is a classifier on the four output-level signals. We fit a logistic regression, a logistic regression on rank-transformed signals, and gradient-boosted trees, and report the variant with the highest full-population AUROC in each condition, a choice that favors this baseline. The \emph{pre-generation probe} is the same linear probe at the same layer, read at the last prompt token before any answer token exists \citep{miftakhova2026pregen}. The \emph{probe+output} predictor \textit{stacks} the probe and the four output-level signals.

\subsection{Evaluation protocol and validity controls}\label{sec:eval}
The evaluation is based on
5-fold cross-validation with \textbf{document-grouped, class-stratified folds} on both benchmarks.
TAT-QA asks 1{,}138 questions over 278 source
documents, so a naive random split on questions would place the same document in both the training and testing samples. This leakage inflates AUROC by up to $0.03$ by our measurements. Thus, all questions on a source document are assigned to the same fold.

Every example is scored out-of-fold, and no model is ever used to predict anything in its own training data. When signals are stacked into a combined predictor, an
inner cross-validation loop within each training fold produces the first-stage scores, so
the second-stage combiner never trains on its own out-of-fold inputs.

We report two performance metrics. The primary metric is AUROC, which measures how well a binary classifier separates wrong from correct answers across all decision thresholds. The secondary metric is PR-AUC, the area under the precision--recall curve (wrong answer = positive). All PR metrics and population shares are importance-reweighted to population prevalence. Uncertainty is reported as 95\% intervals from a paired bootstrap with 2{,}000 replicates. The bootstrap resamples source documents, so all questions on a document move together, and both signals of a comparison are scored on the same resample.

Every reported effect is read against the following controls: (i) a
\emph{permuted-label control} ($\approx$0.5 observed as expected),
(ii) a \emph{random-initialization placebo}, in which the same linear probe is trained on activations from a randomly initialized copy of Qwen3-8B (800-question FinQA pilot, question-level folds) and still reaches 0.625 AUROC by exploiting surface features of the text, so gains from the trained model should be read against $\approx$0.63 rather than against chance, (iii) a \emph{layer-0 baseline} capturing non-contextual token information
($\approx$0.62--0.67), (iv) \emph{cross-task transfer evaluation}
(\S\ref{sec:shift}), and (v) \emph{cross-run reproducibility}, under which activations re-extracted for the same inputs must match across runs.

Fold assignment and subsampling use fixed seeds.

\section{Results}\label{sec:results}

\subsection{Detection performance}\label{sec:fullpop}

\begin{table*}[t]\centering\small
\caption{\textbf{Detection over all answers and among unsure answers ($c(x)<1$).} AUROC. Logprob = best token log-probability score; single = best single output-level signal; comb.\ = supervised combination of the four output-level signals; pre-gen.\ = pre-generation probe; probe+out.\ = probe stacked with the output-level signals. Bold = best single deployable signal per block.
$\dagger$\,Self-consistency requires $k{=}8$ regenerations per query and is reported for reference only.}
\label{tab:fullpop}
\setlength{\tabcolsep}{4.5pt}
\begin{tabular}{llrrrrrrrrrrr}
\toprule
& & \multicolumn{7}{c}{all answers} & \multicolumn{4}{c}{unsure answers}\\
\cmidrule(lr){3-9}\cmidrule(lr){10-13}
& & logprob & \ptrue{} & comb. & self-cons.$^\dagger$ & pre-gen. & probe & probe+out. & single & comb. & pre-gen. & probe\\
\midrule
\multirow{3}{*}{FinQA}
& Qwen3-8B  & .605 & .692 & .705 & .651 & .723 & \textbf{.786} & .794 & .669 & .655 & .620 & \textbf{.722}\\
& Llama-3.1-8B & .705 & .703 & .749 & .820 & .684 & \textbf{.764} & .792 & .680 & .711 & .640 & \textbf{.730}\\
& Gemma-2-9B & .651 & .657 & .724 & .724 & .720 & \textbf{.776} & .796 & .651 & .692 & .626 & \textbf{.715}\\
\midrule
\multirow{3}{*}{TAT-QA}
& Qwen3-8B  & .649 & .725 & \textbf{.735} & .689 & .653 & .732 & .752 & .691 & \textbf{.715} & .683 & .697\\
& Llama-3.1-8B & .654 & .726 & \textbf{.758} & .839 & .648 & .751 & .781 & .679 & \textbf{.692} & .558 & .658\\
& Gemma-2-9B & .588 & .696 & .704 & .770 & .660 & \textbf{.733} & .714 & .613 & \textbf{.621} & .468 & .543\\
\bottomrule
\end{tabular}
\end{table*}

\begin{table}[t]\centering\small
\caption{\textbf{Detection among confident answers (headline).} AUROC among unanimously
confident answers (8/8 self-consistent). Columns as in Table~\ref{tab:fullpop}.
$\Delta$ = probe's gain over the best single signal, with a 95\% paired document-clustered bootstrap interval.}
\label{tab:headline}
\setlength{\tabcolsep}{1.9pt}
\begin{tabular}{lrccccl}
\toprule
 & $n$ (wrong) & single & comb. & pre-gen. & probe & \multicolumn{1}{c}{$\Delta$ [95\% CI]}\\
\midrule
\multicolumn{7}{l}{\emph{FinQA}}\\
Qwen3  & 2{,}503 (1{,}137) & .615 & .646 & .727 & \textbf{.766} & $+.151$ {\footnotesize[.125, .178]}\\
Llama-3.1 & 1{,}085 (329)  & .550 & .540 & .699 & \textbf{.701} & $+.151$ {\footnotesize[.098, .201]}\\
Gemma-2 & 2{,}141 (861)    & .624 & .646 & .715 & \textbf{.755} & $+.131$ {\footnotesize[.101, .160]}\\
\midrule
\multicolumn{7}{l}{\emph{TAT-QA}}\\
Qwen3  & 459 (74)          & .629 & .642 & .592 & \textbf{.687} & $+.058$ {\footnotesize[$-$.034, .118]}\\
Llama-3.1 & 307 (44)       & .587 & .617 & .613 & \textbf{.708} & $+.121$ {\footnotesize[$-$.011, .205]}\\
Gemma-2 & 434 (72)         & .593 & .602 & .640 & \textbf{.684} & $+.091$ {\footnotesize[.002, .164]}\\
\bottomrule
\end{tabular}
\end{table}

\paragraph{Full sample.} Table~\ref{tab:fullpop} shows that, across model families and
datasets, the probe attains 0.73--0.79 AUROC, against 0.66--0.73 for \ptrue{} and
0.59--0.71 for the best token log-probability variant. The supervised output combination reaches 0.70--0.76. On FinQA the probe leads the output combination by $+0.08$ and $+0.05$ for Qwen3-8B and Gemma-2-9B, with intervals that exclude zero, and by $+0.02$ [0.00, 0.03] for Llama-3.1-8B. On TAT-QA the two are statistically indistinguishable in all three families. The answer-conditioned probe also leads the pre-generation probe in all six conditions, by $+0.06$ to $+0.10$. Stacking the output-level signals onto the probe
increases AUROC in five out of six conditions, with the largest gain for Llama-3.1-8B, whose
outputs are the most informative. Once the \textit{non-deployable} self-consistency is
included, the ranking depends on the specific model-benchmark combination.

\paragraph{Confident answers.} Table~\ref{tab:headline} compares the signals among answers classified as \textit{confident}.
Among these answers the output-level signals decay toward the chance level.
This is expected, since membership in the confident group means the LLM itself reproduces the answer under resampling. The probe, in
contrast, barely degrades from its full-population level. On FinQA the probe leads the best single output-level signal by $\Delta =
+0.151, +0.151, +0.131$ AUROC across the three families, and every interval excludes zero. The lead over the supervised output combination is $+0.120$, $+0.161$, and $+0.109$, again with intervals that exclude zero. In PR-AUC terms the probe reaches 0.56, 0.37, and 0.50 against 0.35, 0.18, and 0.29 for the best single signal. At the looser $\ge$6/8 confidence threshold (untabulated), the probe's gains on FinQA range from $+0.11$ to $+0.13$.
On TAT-QA the point estimates are $+0.06$ to $+0.12$. However, the confident group contains only 44--74 wrong answers, and two of the three intervals include zero, so the TAT-QA advantage is not established at this sample size.

The pre-generation probe recovers most of the advantage among confident FinQA answers (0.70--0.73 AUROC). The pre-generation probe leads the supervised output combination by $+0.07$ to $+0.16$, with intervals that exclude zero. The answer-conditioned probe leads the pre-generation probe by $+0.039$ [0.017, 0.061] for Qwen3-8B and $+0.040$ [0.014, 0.065] for Gemma-2-9B, and by $+0.003$ [$-0.035$, 0.041] for Llama-3.1-8B. Hence, among confident answers, internal states read before generation already separate wrong from correct answers better than output-level signals do, and conditioning on the answer adds a further, smaller gain for the two over-committing models.

Additional analyses support the same conclusion.
First, probe score correlates with correctness at $r = 0.23$--$0.38$ on FinQA \emph{after} linearly removing every confidence measure.
Second, the probe--self-consistency rank correlation is only 0.28--0.48.
Third, the advantage is not due to the probe flagging refusals. Abstentions, such as ``the filing does not provide this figure,'' account for 4--11\% of the confident-wrong cell on FinQA. Removing them changes the gap between the probe and the best single signal by about 0.01 in every family (to $+0.161$, $+0.156$, and $+0.120$).

\begin{figure*}[t]\centering
\includegraphics[width=0.86\linewidth]{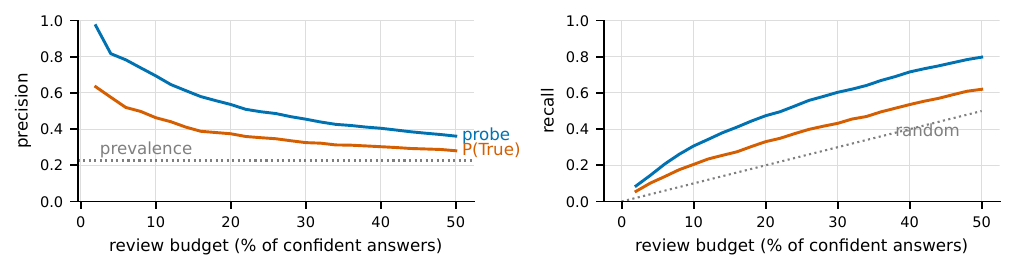}
\caption{Precision (left) and recall (right) for retrieving confident-wrong answers
vs.\ review budget, among unanimously confident answers (Qwen3-8B, FinQA;
population-reweighted). Dotted: prevalence / random routing.}
\label{fig:operating}
\end{figure*}

Figure~\ref{fig:operating} translates the rankings produced by \ptrue{} and the probe into a review policy among confident answers (Qwen3-8B on FinQA). The review budget is the fraction of that pool inspected, most suspicious first. The probe stays above \ptrue{} at every budget in both precision and recall. Flagging the 5\% most-suspicious answers yields 79\% precision (3.4$\times$ prevalence) at 17\% recall for the probe, and a 30\% budget yields 45\% precision at 60\% recall.

\paragraph{Unsure answers.} The right block of Table~\ref{tab:fullpop} reports the comparison within unsure answers.
Here the output-level signals retain
discriminative power. On FinQA the probe leads the best single signal by $+0.05$ to $+0.06$. The probe's lead over the supervised output combination is $+0.07$ for Qwen3-8B, and $+0.02$ for Llama-3.1-8B and Gemma-2-9B with intervals that include zero. On TAT-QA the output combination is ahead in all three families. Thus, the probe does not
dominate output-level signals in all situations.
The probe's value is strongest among confident answers, where errors are hard to detect from output-level signals.

\subsection{Which errors the probe detects}\label{sec:errtypes}

\begin{table}[t]\centering\small
\caption{\textbf{Error types in the confident-wrong cell (FinQA).} Share of the cell, and AUROC for separating each type from confident-correct answers, for Qwen3\,/\,Llama-3.1\,/\,Gemma-2. Output comb.\ = supervised output combination. A dash marks fewer than 20 answers.}
\label{tab:errtypes}
\setlength{\tabcolsep}{3pt}
\begin{tabular}{lccc}
\toprule
type & share (\%) & probe & output comb.\\
\midrule
computation & 38\,/\,53\,/\,38 & .73\,/\,.70\,/\,.74 & .56\,/\,.50\,/\,.56\\
selection   & 38\,/\,27\,/\,35 & .76\,/\,.66\,/\,.73 & .68\,/\,.57\,/\,.69\\
sign        & 11\,/\,12\,/\,12 & .76\,/\,.74\,/\,.81 & .55\,/\,.47\,/\,.61\\
scale       & 4\,/\,3\,/\,4    & .72\,/\,--\,/\,.73  & .73\,/\,--\,/\,.68\\
abstention  & 9\,/\,4\,/\,10   & .98\,/\,--\,/\,.83  & .96\,/\,--\,/\,.84\\
\bottomrule
\end{tabular}
\end{table}

To locate the source of the advantage, we type every wrong FinQA answer mechanically, using the gold reasoning program and table of the original FinQA release \citep{chen2021finqa}. A \emph{sign} error matches the gold answer in magnitude but not in sign. A \emph{scale} error differs from the gold answer by a power of ten that the grader does not forgive. A \emph{selection} error omits at least one operand of the gold program from the model's written reasoning, as when the model reads a different reporting item or period. A \emph{computation} error contains every gold operand in the reasoning but reports a wrong value, as when the model applies a wrong operation or makes an arithmetic slip.

Table~\ref{tab:errtypes} reports the composition of the confident-wrong cell and the detection performance by type. Computation errors are the largest type (38--53\% of the cell), followed by selection errors (27--38\%). The probe's advantage is largest for computation and sign errors. On these two types the supervised output combination reaches only 0.47--0.61, whereas the probe reaches 0.70--0.81. The advantage is smaller for selection errors, which the output-level signals detect at 0.57--0.69. The probe shows no consistent advantage for scale errors, and all signals detect abstentions well. The typing is mechanical. The typing does not separate a wrong period from a wrong reporting item within selection errors, and it is not available for TAT-QA, which is distributed without reasoning programs. We therefore do not attribute the smaller TAT-QA gains or the transfer failure to specific error types.

\subsection{Probe design across layers and pooling}\label{sec:design}

\begin{figure*}[t]\centering
\includegraphics[width=0.84\textwidth]{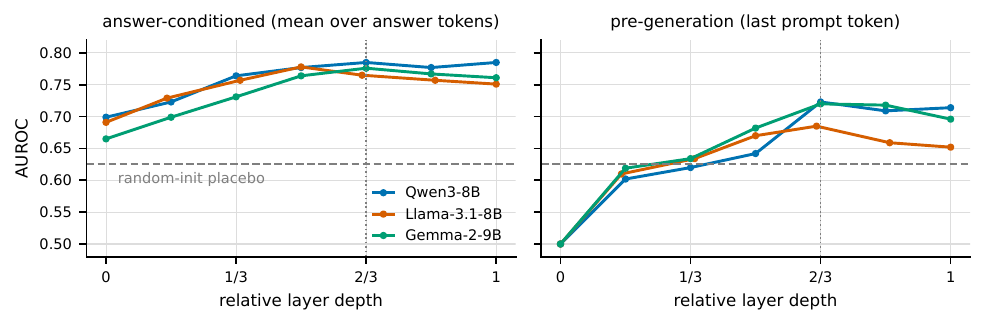}
\caption{Probe AUROC by relative network depth (FinQA). Left:
answer-conditioned probes (mean over generated tokens). Right: pre-generation probes
(last prompt token), which see only the question and its context. Dashed:
random-initialization placebo (0.625). Dotted vertical: the $2/3$-depth layer used
throughout.}
\label{fig:depth}
\end{figure*}

\begin{table}[t]\centering\small
\caption{\textbf{Errors caught at fixed reviewer capacity (FinQA).} Share of all wrong answers
caught within the reviewed fraction, by routing signal. CW rows give the share of confident-wrong answers caught at a 20\% budget. No policy uses the eight-sample confidence label. Bold = most errors caught per row.}
\label{tab:econ}
\setlength{\tabcolsep}{3pt}\small
\begin{tabular}{llrrrrr}
\toprule
& capacity & \ptrue{} & comb. & pre-gen. & probe & two-stage\\
\midrule
\multirow{4}{*}{Qwen3}
& 10\% & 22\% & 22\% & 22\% & \textbf{27\%} & \textbf{27\%}\\
& 20\% & 37\% & 38\% & 38\% & \textbf{45\%} & \textbf{45\%}\\
& 30\% & 50\% & 51\% & 52\% & \textbf{58\%} & \textbf{58\%}\\
& CW, 20\% & 23\% & 25\% & 38\% & \textbf{40\%} & 38\%\\
\midrule
\multirow{4}{*}{Llama-3.1}
& 10\% & 21\% & 22\% & 18\% & 22\% & \textbf{23\%}\\
& 20\% & 36\% & 38\% & 33\% & 38\% & \textbf{40\%}\\
& 30\% & 48\% & 51\% & 45\% & 51\% & \textbf{54\%}\\
& CW, 20\% & 7\% & 11\% & \textbf{30\%} & 26\% & 16\%\\
\midrule
\multirow{4}{*}{Gemma-2}
& 10\% & 18\% & 21\% & 22\% & \textbf{26\%} & \textbf{26\%}\\
& 20\% & 32\% & 38\% & 39\% & 44\% & \textbf{45\%}\\
& 30\% & 44\% & 53\% & 52\% & 58\% & \textbf{60\%}\\
& CW, 20\% & 22\% & 22\% & \textbf{34\%} & 33\% & \textbf{34\%}\\
\bottomrule
\end{tabular}
\end{table}

Figure~\ref{fig:depth} reports the layer-by-pooling sweep behind the fixed design, and
two regularities hold across all three families. First, mean pooling over all generated answer tokens (left panel) has higher AUROC than reading the last prompt token (right panel). Second, the layer profile is not monotonic. In the left panel, AUROC rises from the embedding layer (0.67--0.70) to approximately 2/3 of the depth and plateaus there until the final layer (0.76--0.79).

Hence, the a priori choice of a mean-pooled probe at two-thirds depth is close to optimal and
transfers across LLM families. For Qwen3-8B and Gemma-2-9B the two-thirds-depth layer coincides
with the sweep-optimal layer (0.785 and 0.776 AUROC). For Llama-3.1-8B the optimum sits
earlier, near one-half depth (0.778 at layer 16), and fixing the layer at two-thirds depth costs only 0.013 AUROC (0.765 at layer 21).

\label{sec:shift}Cross-dataset transfer asks whether a probe trained on one benchmark still works on the other. Such a probe does not. Training on FinQA and testing on TAT-QA, or the reverse, costs roughly 0.07--0.16 AUROC across all three families, and the transferred probe typically falls below the target's own output-level signals. The hallucination probe therefore appears task-specific, and it must be trained for new tasks.

\subsection{Triage under bounded human review}\label{sec:econ}
Capacity for human review is limited in financial LLM applications, so a triage system must direct the answers that are most likely wrong to reviewers. We evaluate routing policies that a deployed system can execute. None of the policies uses the eight-sample confidence label. Each single-signal policy reviews the answers with the lowest scores under that signal. The \emph{two-stage} policy first spends a fraction $\alpha$ of the budget on the answers with the lowest output-combination scores, which are the visibly uncertain answers, and fills the remainder by probe score. The fraction $\alpha$ is chosen on one document-grouped half of the data, and all results are reported on the other half.

Table~\ref{tab:econ} reports the share of errors caught at review budgets of 10--30\% of answers. For the over-committing models, probe routing catches more errors than every output-level policy at every budget. At a 20\% budget the probe's gain over \ptrue{} is $+7.5$ points [3.4, 11.6] for Qwen3-8B and $+12.0$ points [7.6, 16.6] for Gemma-2-9B, and its gain over the output combination is $+6.7$ and $+6.1$ points, with intervals that exclude zero. For Llama-3.1-8B, whose outputs are already informative, the probe matches the output combination. The two-stage policy does not reliably improve on probe routing alone. The two-stage gain at a 20\% budget is $+0.4$, $+1.8$, and $+0.4$ points, with intervals that include zero, and the largest gain is $+2.4$ points [0.2, 4.7] for Llama-3.1-8B at a 30\% budget. A single ranking by probe score is therefore the simplest executable policy.

The CW rows show where the signals differ. The output-level policies catch 7--25\% of confident-wrong answers at a 20\% budget, whereas the two internal-state probes catch 26--40\%. The two-stage policy catches fewer confident errors than probe routing for Llama-3.1-8B (16\% against 26\%), where the chosen $\alpha$ assigns 80\% of the budget to the output combination. On TAT-QA the evaluation half contains only 70--146 wrong answers. At a 20\% budget the probe's gain over \ptrue{} is 0 to $+7$ points there, with intervals of about $\pm$12 points.

The overhead of the policies also differs. The probe and the pre-generation probe add one inner product on a 3{,}584- or 4{,}096-dimensional vector per answer. \ptrue{}, and any policy that uses the output combination, adds a second forward pass over a sequence about as long as the original prompt and answer. Self-consistency adds eight generations.

\section{Discussion and limitations}\label{sec:limits}
In a human-in-the-loop deployment, the results in this paper support a monitor that ranks answers for review by probe score. The probe's value is largest among confident answers, where output-level signals carry little information, and for computation and sign errors. For an institution, such a monitor is an automated
control in the oversight of its generative LLM systems, one that does not certify answers but helps prioritize them for review. Several limitations bound these claims.

First, the probe requires direct access to activations inside the LLM. Thus, it is only deployable for self-hosted LLMs rather than third-party APIs. In the latter case, \ptrue{} and token log-probabilities remain the only viable options.

Second, our evaluation protocol is limited on purpose to remain tractable, which affects the study's generalizability. The experiments are restricted to numerical answers, which can be graded mechanically, and the findings may not extend to free-text financial QA. The protocol covers prompted CoT rather than native reasoning modes, and single-turn QA rather than multi-turn dialogues. We test three LLMs in the 8B--9B parameter range, and larger models are not tested. Our baselines are output-level confidence signals, and we do not evaluate detectors that analyze the generated text or reasoning trace.

Third, we make no claim about robustness to temporal drift in accounting or reporting
practice, because the arithmetic conventions that FinQA questions exercise have been stable across its filing years.

\section{Conclusion}
This paper asked how much hallucination signal remains recoverable from an
LLM's internal states when its output-level signals convey confidence. We compared a linear probe on the
residual stream against the output-level signals a deployed system could afford, on two numerical financial QA benchmarks and three model
families.

Three findings emerged. \textit{First}, confident hallucination is common in numerical financial QA.
Depending on the LLM family, 5--19\% of all FinQA answers are unanimously self-consistent
and wrong. \textit{Second}, among those confident answers the output-level signals approach chance level, while the probe retains 0.68--0.77 AUROC. The advantage of $+0.13$ to $+0.15$ AUROC on FinQA
is replicated in all three families, holds against a supervised combination of output-level signals, and is largest for computation and sign errors. However, it does not survive task transfer, so a probe-based monitor must be retrained per task. \textit{Third}, probe-ranked routing catches more errors than output-level routing at realistic review capacities for the two over-committing models. Extending the analysis to
reasoning-mode traces, multi-turn conversations, free-text answers, and larger LLMs remains for future work.

\section*{Code and data availability}
The code, the exact prompts, the decoding configuration, the grading and parsing rules, the model identifiers, and per-example score files that reproduce every reported number are available at \url{\codeurl}.

\bibliographystyle{ACM-Reference-Format}
\bibliography{references}

\end{document}